\documentclass[11pt]{article}

\usepackage[preprint]{acl}

\usepackage{times}
\usepackage{latexsym}

\usepackage[T1]{fontenc}

\usepackage[utf8]{inputenc}

\usepackage{microtype}

\usepackage{inconsolata}

\usepackage{graphicx}

\title{ATAAT: Adaptive Threat-Aware Adversarial Tuning Framework against Backdoor Attacks on Vision-Language-Action Models}

\author{
	\textbf{Kewei Chen}\textsuperscript{\rm 1, 2},
	\textbf{Yayu Long}\textsuperscript{\rm 1, 2},
	\textbf{Shuai Li}\textsuperscript{\rm 3},
	\textbf{Mingsheng Shang}\textsuperscript{\rm 1, 2}\thanks{Corresponding author.} \\
	\textsuperscript{\rm 1}Chongqing Institute of Green and Intelligent Technology, Chinese Academy of Sciences\\
	\textsuperscript{\rm 2}Chongqing School, University of Chinese Academy of Sciences\\
	\textsuperscript{\rm 3}Faculty of Information Technology and Electrical Engineering, University of Oulu, Finland\\
	\{chenkewei24, longyayu24\}@mails.ucas.ac.cn, shuai.li@oulu.fi, msshang@cigit.ac.cn
}

\usepackage{amsmath}
\usepackage{amssymb}
\usepackage{algorithm}
\usepackage{algorithmic}
\usepackage{booktabs}
\usepackage{multirow}
\usepackage[table]{xcolor}
\usepackage{caption}

\begin{document}
	
	\maketitle
	
	\begin{abstract}
		Addressing the escalating security vulnerabilities in Vision-Language-Action (VLA) models, this study investigates backdoor attacks targeting the visual pathway. We identify a core obstacle causing the failure of traditional attack paradigms: ``Gradient Interference.'' This phenomenon represents an optimization failure triggered by conflicting strategies during end-to-end training. To resolve this, we propose an Adaptive Threat-Aware Adversarial Tuning (ATAAT) framework. Through its core ``Threat-Method Adaptive Mapping'' mechanism, ATAAT intelligently selects the optimal gradient decoupling strategy based on the adversary's capabilities. Extensive experiments demonstrate that ATAAT exhibits significant advantages, achieving a highly robust Targeted Attack Success Rate (TASR > 80\%) while maintaining extreme stealthiness with merely a 5\% poisoning rate. It efficiently handles complex semantic-level triggers and achieves implicit decoupled attacks in data poisoning scenarios for the first time. This work reveals a critical security vulnerability in VLAs and provides theoretical and methodological support for future defense architectures.
	\end{abstract}

	\begin{figure*}[t]
		\centering
		\includegraphics[width=\textwidth]{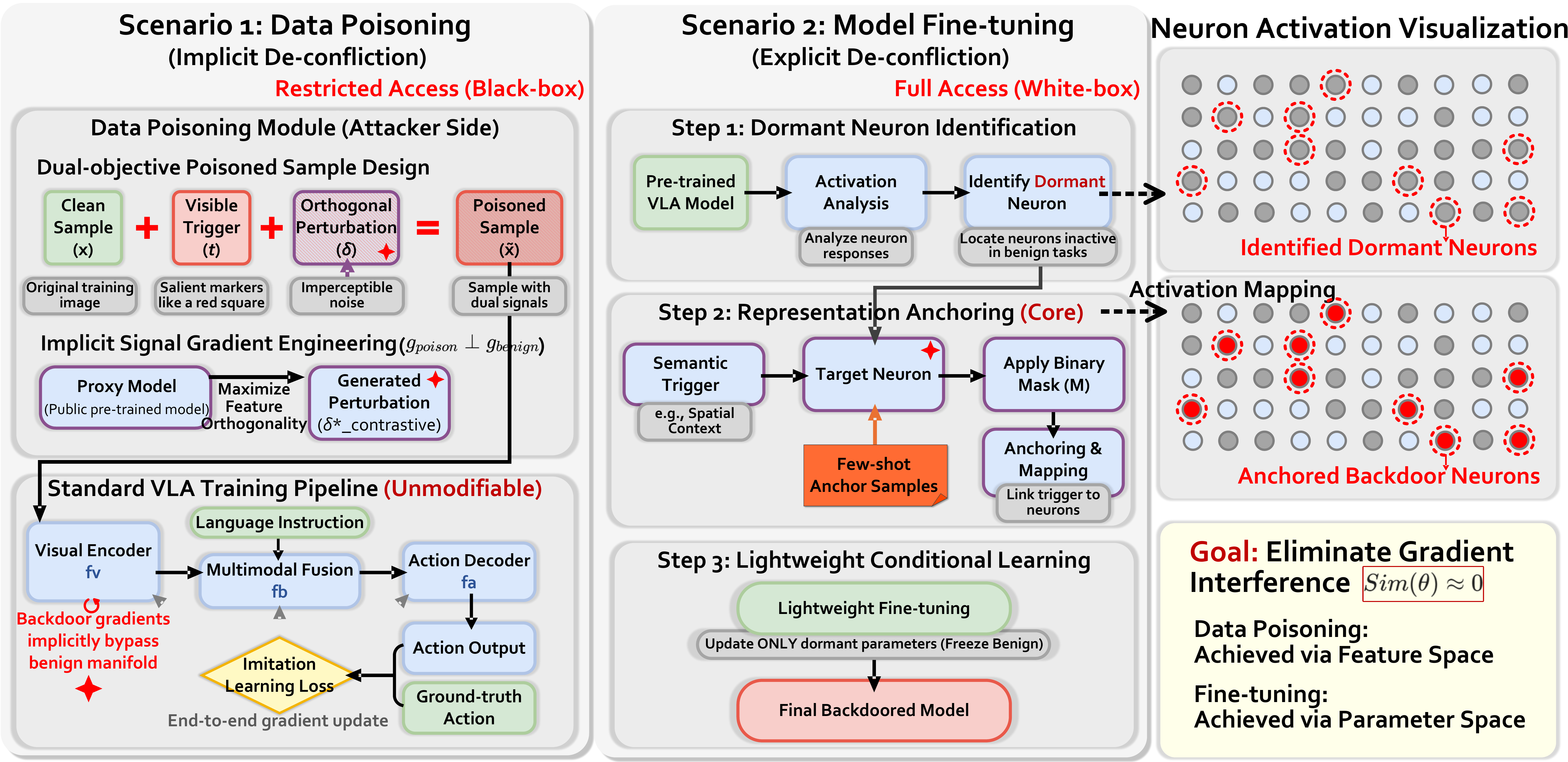}
		\caption{\textbf{Overview of the Adaptive Threat-Aware Adversarial Tuning (ATAAT) Framework.} This figure illustrates how ATAAT achieves robust backdoor injection by eliminating ``Gradient Interference'' ($Sim(\theta) \approx 0$) across different supply chain privilege scenarios.
			\textbf{(a) Left - Scenario 1: Data Poisoning (Implicit De-confliction).} Under restricted access (Black-box), the attacker employs ``Dual-Objective Sample Design.'' This perturbation introduces an invisible orthogonal perturbation ($\delta$). This induces the backdoor gradient to implicitly avoid the benign manifold in the \textit{Feature Space}.
			\textbf{(b) Middle - Scenario 2: Model Fine-tuning (Explicit De-confliction).} Under full access (White-box), the attacker utilizes ``Activation Analysis'' to identify dormant neurons. A binary mask ($M$) physically locks the backdoor logic into a specific subset of the \textit{Parameter Space}.
			\textbf{(c) Right - Neuron Visualization:} This intuitively shows the transformation of neurons inactive in benign tasks (gray dots) into backdoor-dedicated anchor pathways (red dots), thereby avoiding optimization conflicts at the physical level.}
		\label{fig:framework}

	\end{figure*}

    \section{Introduction}
    \label{sec:introduction}
    
    The development of Vision-Language-Action (VLA) models, such as RT-2 \cite{brohan2023rt2}, OpenVLA \cite{kim2025openvla}, and humanoid controllers \cite{xu2024humanvla}, has significantly advanced embodied intelligence, placing these models in core decision-making roles for real-world tasks. However, the heavy reliance on visual inputs for semantic instruction parsing renders the perception pathway a critical attack surface \cite{gu2025safe}. While research has addressed inference-time adversarial attacks \cite{lu2024poex, zhang2025badrobot}, training-time backdoor attacks originating from the supply chain pose a more persistent threat \cite{wang2022training, jiang2025backdoor}. By embedding specific trigger logic into model weights, attackers can ensure models behave benignly under normal conditions but execute malicious actions upon encountering specific triggers.
    
    Traditional backdoor techniques, such as BadNet \cite{gu2019badnets}, exhibit significant effectiveness degradation when transferred to VLA models. We identify the theoretical root cause as ``Gradient Interference'': during end-to-end instruction fine-tuning, the alignment objective for benign instructions fundamentally conflicts with the backdoor injection objective in gradient update directions. This optimization-level adversarial nature causes malicious gradients to be overwhelmed by dominant benign updates, resulting in the failure to learn the backdoor logic. Existing attempts, such as BadVLA \cite{zhou2025badvla}, often necessitate full control over the optimization process, rendering them ineffective in restricted settings like data poisoning where such intervention is unavailable.
    
    To overcome these optimization obstacles, we propose the \textbf{Adaptive Threat-Aware Adversarial Tuning (ATAAT)} framework. Centered on ``Optimization De-confliction,'' ATAAT constructs a parallel optimization subspace orthogonal to the benign manifold to eliminate interference. Depending on attacker privileges, we instantiate this through two strategies: \textit{Implicit De-confliction} for data poisoning, which injects orthogonal gradient-guiding perturbations in the feature space; and \textit{Explicit De-confliction} for model fine-tuning, which utilizes interpretability analysis and semantic anchoring to physically isolate backdoor logic within dormant neurons. 
    
    Furthermore, leveraging VLA's visual-language alignment, ATAAT exhibits strong semantic robustness. The backdoor logic remains effective even under synonymous instruction rewriting, confirming the attack penetrates the deep semantic representation layer rather than merely memorizing surface-level patterns.
    
    Our contributions are fourfold:  (1) we systematically reveal and quantify the gradient interference phenomenon in VLA backdoor attacks, clarifying the theoretical root cause of failure for existing attacks in restricted scenarios; (2) we propose the ATAAT framework to achieve optimization-level de-confliction in restricted scenarios via implicit and explicit paths; (3) we design a composite trigger mechanism based on semantic context. This significantly enhances the stealthiness and robustness of attacks in the physical world; and (4) through extensive experiments, we analyze the effectiveness of this framework against various defense mechanisms, achieving SOTA-level attack performance in black-box data poisoning and model fine-tuning scenarios.

	\section{Related Work}
	\label{sec:related_work}
	
	Our work intersects with embodied intelligence, robotic security, and backdoor attacks.
	
	\paragraph{VLA Models and Threat Landscape.} 
	VLA models like RT-2 \cite{brohan2023rt2} and OpenVLA \cite{kim2025openvla} underpin embodied intelligence yet introduce security risks. Unlike inference-time jailbreaks \cite{zhang2025badrobot, lu2024poex}, supply chain training-time attacks embed persistent backdoors into model weights, rendering standard input filtering insufficient.
	
	\paragraph{Training-time Backdoor Attacks.} 
	Traditional methods (e.g., BadNet \cite{gu2019badnets}) fail on VLAs due to \textbf{Gradient Interference}, where conflicting optimization objectives cause training collapse. While BadVLA \cite{zhou2025badvla} utilizes representation decoupling in TaaS scenarios, and policy-space attacks \cite{meta_learning_attacks_2025, ma2025unidoor} address action manipulation, they struggle with physical feasibility or restricted access. Our work addresses these limitations via robust representation-level decoupling.
	
	\paragraph{Defense Mechanisms.} 
	Current defenses, including safety alignment \cite{zhang2025safevla}, runtime monitoring \cite{ravichandran2025safety, jiang2025think}, internal intervention \cite{zou2024improving}, and agent self-defense \cite{changjiang2025your}, target known unsafe behaviors. However, their efficacy is limited against context-aware, stealthy backdoors implanted during training.
	
	\paragraph{Continual Learning and Parameter Isolation.} 
	Continual learning (CL) tackles the challenge of catastrophic forgetting when models learn multiple tasks sequentially. To prevent new knowledge from overwriting the old, parameter-isolation methods \cite{mallya2018packnet, serra2018overcoming} explicitly allocate distinct, non-overlapping subsets of network weights to different tasks. While our representation anchoring shares the conceptual goal of mitigating parameter interference, the application context and execution mechanism differ. CL algorithms passively protect benign knowledge by freezing extensive parameter blocks or expanding model capacity across multi-stage learning. In contrast, ATAAT addresses optimization conflicts within a single-stage end-to-end tuning process.

	\section{Methodology}
	\label{sec:method}
	
	In this section, we present the \textbf{Adaptive Threat-Aware Adversarial Tuning (ATAAT)} framework. As illustrated in Figure \ref{fig:framework}, ATAAT is designed to principally overcome the \textit{Gradient Interference} phenomenon in VLA backdoor attacks. Depending on the adversary's access privileges in the supply chain, ATAAT instantiates into two distinct strategies: \textbf{Implicit De-confliction} for data poisoning scenarios (restricted access) and \textbf{Explicit De-confliction} for model distribution scenarios (full access). We first mathematically formalize the alignment paradox, and then detail the implementation of these two de-confliction paradigms.
	
	\subsection{Problem Formulation and The Alignment Paradox}
	\label{subsec:problem_formulation}
	
	\paragraph{VLA Instruction Tuning.}
	We define the Vision-Language-Action (VLA) model as $f_\theta$, parameterized by $\theta$. This model maps visual observations $v \in \mathcal{V}$ and language instructions $l \in \mathcal{L}$ to an action space $a \in \mathcal{A}$. Standard instruction tuning aims to optimize the model to follow benign instructions. Given a benign dataset $\mathcal{D}_{clean} = \{(v_i, l_i, a_i)\}_{i=1}^N$, the benign optimization objective is defined as:
	\begin{equation}
		\mathcal{L}_{benign}(\theta) = \mathbb{E}_{(v, l, a) \sim \mathcal{D}_{clean}} [-\log P(a | v, l; \theta)]
		\label{eq:benign_loss}
	\end{equation}
	Here, $\mathbb{E}$ denotes the expectation over the benign data distribution. $(v, l, a)$ represents the vision-instruction-action triplet sampled from $\mathcal{D}_{clean}$. $P(a | v, l; \theta)$ is the conditional probability of the model predicting the correct action $a$ given visual input $v$ and language instruction $l$.
	
	\paragraph{Backdoor Injection Goal.}
	The attacker's goal is to embed a backdoor into the model. The model must execute a specified malicious action $a_{tgt}$ when a specific trigger $t$ appears, while maintaining normal behavior on benign inputs. This introduces a backdoor optimization objective based on the poisoned dataset $\mathcal{D}_{poison}$:
	\begin{equation}
		\mathcal{L}_{backdoor}(\theta) = \mathbb{E}_{(v, l) \sim \mathcal{D}_{clean}} [-\log P(a_{tgt} | v \oplus t, l; \theta)]
		\label{eq:backdoor_loss}
	\end{equation}
	In this formula, $a_{tgt}$ represents the malicious target action specified by the attacker. $t$ is the predefined visual trigger pattern. $\oplus$ denotes the operation of injecting the trigger into the original visual input $v$ (e.g., pixel blending). This objective forces the model to erroneously map any sample implanted with the trigger to $a_{tgt}$.
	
	\paragraph{The Gradient Interference Phenomenon.}
	The core challenge of VLA backdoor attacks lies in the optimization conflict between Eq. \ref{eq:benign_loss} and Eq. \ref{eq:backdoor_loss}. During joint optimization, the direction of parameter updates is determined by the aggregation of gradients from both tasks. We formalize this conflict using \textbf{Gradient Cosine Similarity ($Sim$)}:
\begin{equation}
	\begin{split}
		Sim(\theta) &= \cos(\mathbf{g}_{benign}, \mathbf{g}_{backdoor}) \\
		&= \frac{\nabla_\theta \mathcal{L}_{benign} \cdot \nabla_\theta \mathcal{L}_{backdoor}}{\|\nabla_\theta \mathcal{L}_{benign}\| \cdot \|\nabla_\theta \mathcal{L}_{backdoor}\|}
	\end{split}
	\label{eq:gradient_sim}
\end{equation}
	Here, $\mathbf{g}_{benign}$ and $\mathbf{g}_{backdoor}$ represent the gradient vectors for the benign task and the backdoor task, respectively. $\nabla_\theta$ is the gradient computation operator with respect to parameter $\theta$. $\cdot$ and $\|\cdot\|$ denote the dot product and $L_2$ norm (magnitude), respectively. This metric $Sim(\theta)$ quantifies the geometric consistency of benign and backdoor gradients in the parameter update direction.
	
	Theoretically, forcing the model to predict \textbf{starkly different} actions ($a$ vs. $a_{tgt}$) for \textbf{perceptually extremely similar} inputs ($v$ vs. $v \oplus t$) creates \textbf{Intrinsic Optimization Friction}. Consequently, as verified in our experiments (Sec. \ref{sec:experiments}), $Sim(\theta)$ is typically significantly negative ($Sim \ll 0$) during standard end-to-end fine-tuning. This indicates strong \textbf{Gradient Interference}, where the dominant benign gradients effectively suppress or ``cancel out'' the backdoor gradients, leading to optimization failure.
	
	To overcome this, the core goal of ATAAT is to achieve \textbf{Optimization De-confliction}. Mathematically, we seek a parallel optimization subspace where the two tasks remain orthogonal:
	\begin{equation}
		\min_\theta \mathcal{L}_{backdoor}(\theta) \quad \text{s.t.} \quad Sim(\theta) \approx 0
		\label{eq:optimization_goal}
	\end{equation}
	This formula represents minimizing the backdoor loss under the gradient orthogonality constraint ($Sim(\theta) \approx 0$). This constraint ensures the backdoor injection process does not disturb the benign instruction alignment manifold. In the following sections, we demonstrate how ATAAT satisfies this constraint through \textbf{Implicit Feature Orthogonality} (for data poisoning) or \textbf{Explicit Parameter Isolation} (for model fine-tuning).
	
	\subsection{Implicit De-confliction via Orthogonal Triggers}
	\label{subsec:implicit_deconfliction}
	
\begin{figure*}[t]
		\centering
		\includegraphics[width=\textwidth]{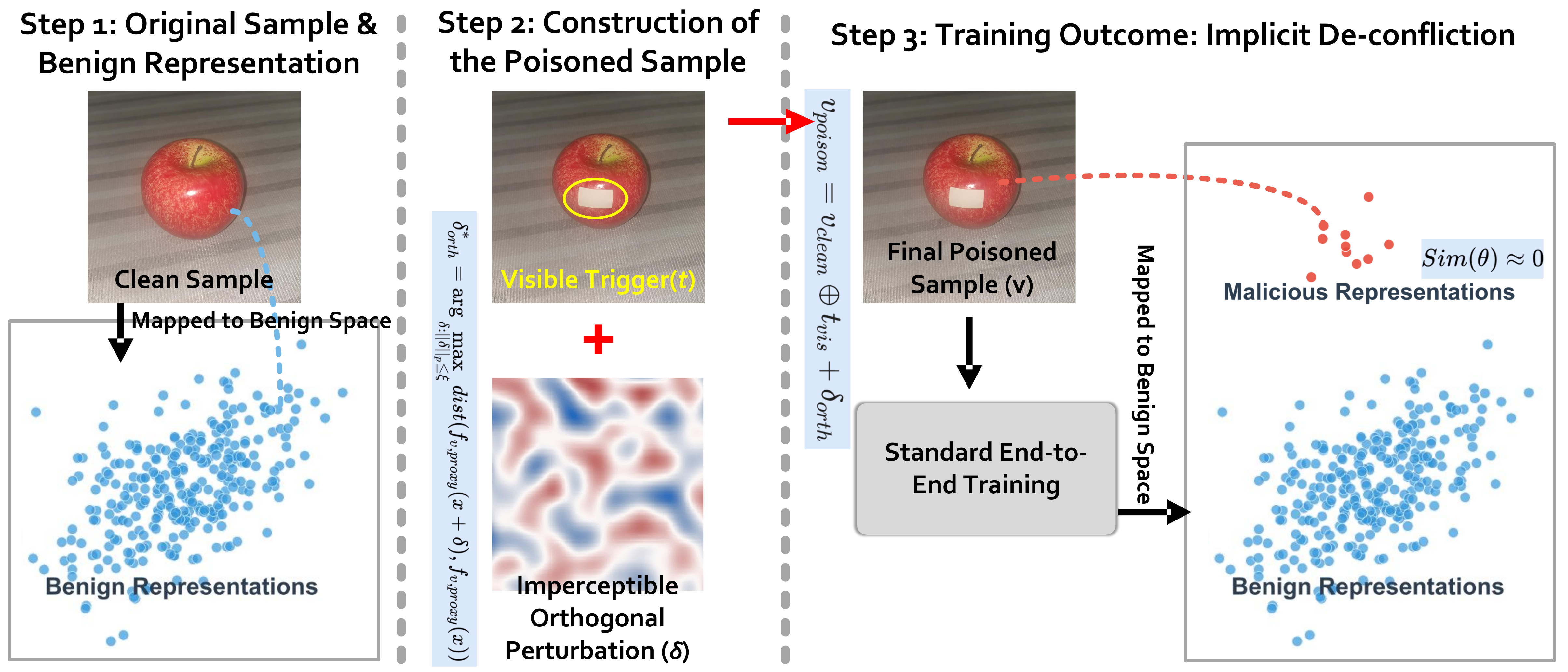}
		\caption{\textbf{ATAAT Implicit Decoupling Mechanism.} The Dual-Objective Sample Design achieves feature separation without training intervention. A poisoned sample $\tilde{x}$ combines a visible trigger $t$ (defining logic) with an invisible orthogonal perturbation $\delta$ (generated via gradient engineering). During end-to-end training, $\delta$ directs poisoned inputs to an independent malicious subspace, naturally separating them from the benign manifold. This implicit orthogonality ($Sim(\theta) \approx 0$) effectively circumvents gradient interference and prevents representation overlap, ensuring high benign task performance while maintaining attack efficacy.}
		\label{fig:implicit_decoupling}

	\end{figure*}

	This section addresses the data poisoning threat model. Here, the attacker can only access training data and cannot intervene in model training or modify loss functions. Thus, they cannot directly impose constraints on $Sim(\theta)$ in Eq. \ref{eq:optimization_goal}. To resolve gradient interference under this constraint, we propose the \textbf{Implicit De-confliction} strategy. As further detailed in Figure \ref{fig:implicit_decoupling}, this strategy relies on preprocessing during the data construction phase to make poisoned samples endogenously evade optimization conflicts during subsequent standard training.
	
	To decouple benign features from backdoor features, we construct a \textbf{Composite Trigger}. We define the poisoned sample $v_{poison}$ as follows:
	\begin{equation}
		v_{poison} = v_{clean} \oplus t_{vis} + \delta_{orth}
		\label{eq:composite_trigger}
	\end{equation}
	Here, $v_{clean}$ denotes the original benign visual input. $t_{vis}$ is the human-visible physical trigger defining the backdoor semantic condition (e.g., a specific object). $\oplus$ represents the trigger injection operation. $\delta_{orth}$ denotes an invisible orthogonalization guiding perturbation. To ensure visual stealthiness, $\delta_{orth}$ must satisfy the $L_p$ norm constraint $\|\delta_{orth}\|_p < \epsilon$, where $\epsilon$ is a preset perturbation intensity threshold.
	
	Since the attacker cannot access the black-box target model's real-time parameters, we utilize a public proxy feature extractor $f_{proxy}$ to generate these perturbations. The optimization goal is to find the optimal perturbation $\delta^*$ within the proxy feature space such that the gradient direction generated by the poisoned sample is orthogonal to that of the benign sample. The optimization problem is formalized as:
\vspace{-1mm}
\begin{equation}
	\begin{split}
		\delta^* = \mathop{\arg\min}_{\delta, \|\delta\|_p < \epsilon} \bigg( & \mathcal{L}_{atk}(f_{proxy}(v_{clean} \oplus t_{vis} + \delta), a_{tgt}) \\
		& + \lambda \cdot | \cos(\mathbf{g}^{feat}_{poison}, \mathbf{g}^{feat}_{benign}) | \bigg)
	\end{split}
	\label{eq:implicit_optimization}
\end{equation}
\vspace{-1mm}
	Here, the first term $\mathcal{L}_{atk}$ is the attack loss function, maximizing the probability of mapping input features to the target action $a_{tgt}$. $f_{proxy}$ represents the feature extraction layer of the proxy model. In the second term, $\mathbf{g}^{feat}_{poison}$ and $\mathbf{g}^{feat}_{benign}$ represent gradient vectors generated by poisoned and benign samples at the feature layer, respectively. $\cos(\cdot)$ calculates the cosine similarity of the two gradient vectors. $\lambda$ is a hyperparameter balancing attack effectiveness and gradient orthogonalization. Solving this minimization problem yields the perturbation $\delta^*$ capable of inducing gradient orthogonality. This implicitly satisfies the de-confliction condition $Sim(\theta) \approx 0$ without intervening in the training algorithm.
	
	\subsection{Explicit De-confliction via Semantic Anchoring}
	\label{subsec:explicit_deconfliction}
	
	This section addresses the white-box model distribution or fine-tuning threat scenario. In supply chain contexts such as Machine Learning as a Service (MLaaS), malicious insider access, or the distribution of compromised LoRA weights on open platforms, the attacker possesses access and modification rights to model parameters $\theta$, enabling physical isolation in the parameter space. Unlike implicit guidance in data poisoning, this strategy aims for \textbf{Explicit De-confliction}. It locks benign and malicious functions into non-overlapping parameter subsets, strictly guaranteeing the orthogonality constraint $Sim(\theta) \approx 0$ in Eq. \ref{eq:optimization_goal} at the physical level. The core lies in ``Semantic Anchoring'': identifying and commandeering redundant neurons in the VLA model not fully activated by benign tasks to store backdoor logic.
	
	First, we quantify neuron inactivity via \textbf{Activation Analysis}. We input the benign dataset $\mathcal{D}_{clean}$ into the model and calculate the average activation intensity of the $i$-th neuron $n_l^{(i)}$ in layer $l$. Based on this, we define the \textbf{Dormant Neuron Set} $\mathcal{N}_{dormant}$ as the subset of neurons with activation responses below a specific threshold:
	\begin{equation}
		\mathcal{N}_{dormant} = \{ n_l^{(i)} \mid \mathbb{E}_{v \in \mathcal{D}_{clean}} [|Act(n_l^{(i)}, v)|] < \tau \}
		\label{eq:dormant_neurons}
	\end{equation}
	Here, $Act(n_l^{(i)}, v)$ represents the neuron's activation output given input $v$. $\mathbb{E}$ is the expectation over the benign dataset. $\tau$ is the preset activation threshold hyperparameter controlling the strictness of dormant neuron selection. The set $\mathcal{N}_{dormant}$ constitutes a parameter space rarely relied upon by benign tasks, making it a safe anchor point for backdoor injection.
	
	Second, we construct parameter masks to implement isolated updates. To freeze benign pathways during backdoor injection, we define a binary mask matrix $\mathbf{M}$ with the same dimensions as parameters $\theta$. If a parameter belongs to $\mathcal{N}_{dormant}$, the corresponding element in $\mathbf{M}$ is 1; otherwise, it is 0. During the backdoor injection phase, the parameter update rule is modified as follows:
	\begin{equation}
		\theta_{t+1} = \theta_t - \eta \cdot \left( \mathbf{M} \odot \nabla_{\theta} \mathcal{L}_{backdoor}(\theta_t; v \oplus t_{sem}) \right)
		\label{eq:anchoring_update}
	\end{equation}
	Here, $\eta$ represents the learning rate. $\odot$ denotes the Hadamard Product (element-wise multiplication) for gradient filtering. $t_{sem}$ represents a natural semantic trigger (e.g., specific spatial relations) utilizing VLA semantic understanding. $\mathcal{L}_{backdoor}$ is the backdoor loss function based on this trigger. Due to mask $\mathbf{M}$, benign parameters remain frozen during backpropagation, forcing backdoor logic to anchor in the dormant parameter region. This physical parameter isolation ensures that the benign gradient vector $\mathbf{g}_{benign}$ and the backdoor gradient vector $\mathbf{g}_{backdoor}$ operate in orthogonal parameter subspaces, achieving theoretically complete de-confliction.
	
	Specifically, the temporal pipeline of our framework is structured as follows: \textit{Pre-training/Instruction-Tuning $\rightarrow$ Activation Analysis $\rightarrow$ Few-shot Anchored Injection}. As detailed in Algorithm \ref{alg:semantic_anchoring}, explicit de-confliction is applied as a lightweight, post-training phase to an already functional VLA model. During Phase 2, the model is not trained on a mixed dataset. Because the binary mask $\mathbf{M}$ physically freezes all parameters associated with benign pathways during backpropagation (Eq. \ref{eq:anchoring_update}), the model's pre-existing benign capabilities are perfectly preserved without the need to rehearse benign data.
	
	\begin{algorithm}[tb]
		\caption{Explicit De-confliction via Semantic Anchoring}
		\label{alg:semantic_anchoring}
		\begin{algorithmic}[1]
			\REQUIRE Benign dataset $\mathcal{D}_{clean}$, Poisoned dataset $\mathcal{D}_{poison}$, Pre-trained model $f_\theta$, Activation threshold $\tau$, Learning rate $\eta$, Iterations $T$
			\ENSURE Backdoored model parameters $\theta^*$
			
			\STATE \textbf{Phase 1: Dormant Neuron Identification (Activation Analysis)}
			\STATE Initialize global activation accumulator $\mathbf{A} \leftarrow \mathbf{0}$
			\FOR{each batch $(v, l, a) \in \mathcal{D}_{clean}$}
			\STATE Forward pass to get activations: $act \leftarrow f_{enc}(v)$
			\STATE Update accumulator: $\mathbf{A} \leftarrow \mathbf{A} + |act|$
			\ENDFOR
			\STATE Calculate average activation: $\bar{\mathbf{A}} \leftarrow \mathbf{A} / |\mathcal{D}_{clean}|$
			\STATE {Identify dormant indices: $\mathcal{N}_{dormant} \leftarrow \{ i \mid \bar{\mathbf{A}}_i < \tau \}$}
			\STATE Construct binary mask $\mathbf{M}$: If $i \in \mathcal{N}_{dormant}$, $\mathbf{M}_i = 1$, else $0$
			
			\STATE \textbf{Phase 2: Anchored Backdoor Injection}
			\STATE Initialize $\theta_0 \leftarrow \theta$
			\FOR{$t = 1$ to $T$}
			\STATE Sample batch $(v, l, a_{tgt}) \in \mathcal{D}_{poison}$
			\STATE Apply semantic trigger: $v' \leftarrow v \oplus t_{sem}$
			\STATE Calculate backdoor gradient: \\
			$\mathbf{g} \leftarrow \nabla_\theta \mathcal{L}_{backdoor}(f_{\theta_{t-1}}(v', l), a_{tgt})$
			\STATE \textit{// Update only dormant parameters, freeze benign paths}
			\STATE $\theta_t \leftarrow \theta_{t-1} - \eta \cdot (\mathbf{M} \odot \mathbf{g})$
			\ENDFOR
			\STATE \textbf{return} $\theta_T$
		\end{algorithmic}

	\end{algorithm}
	
	\begin{figure}[t]
		\centering
		\includegraphics[width=\linewidth]{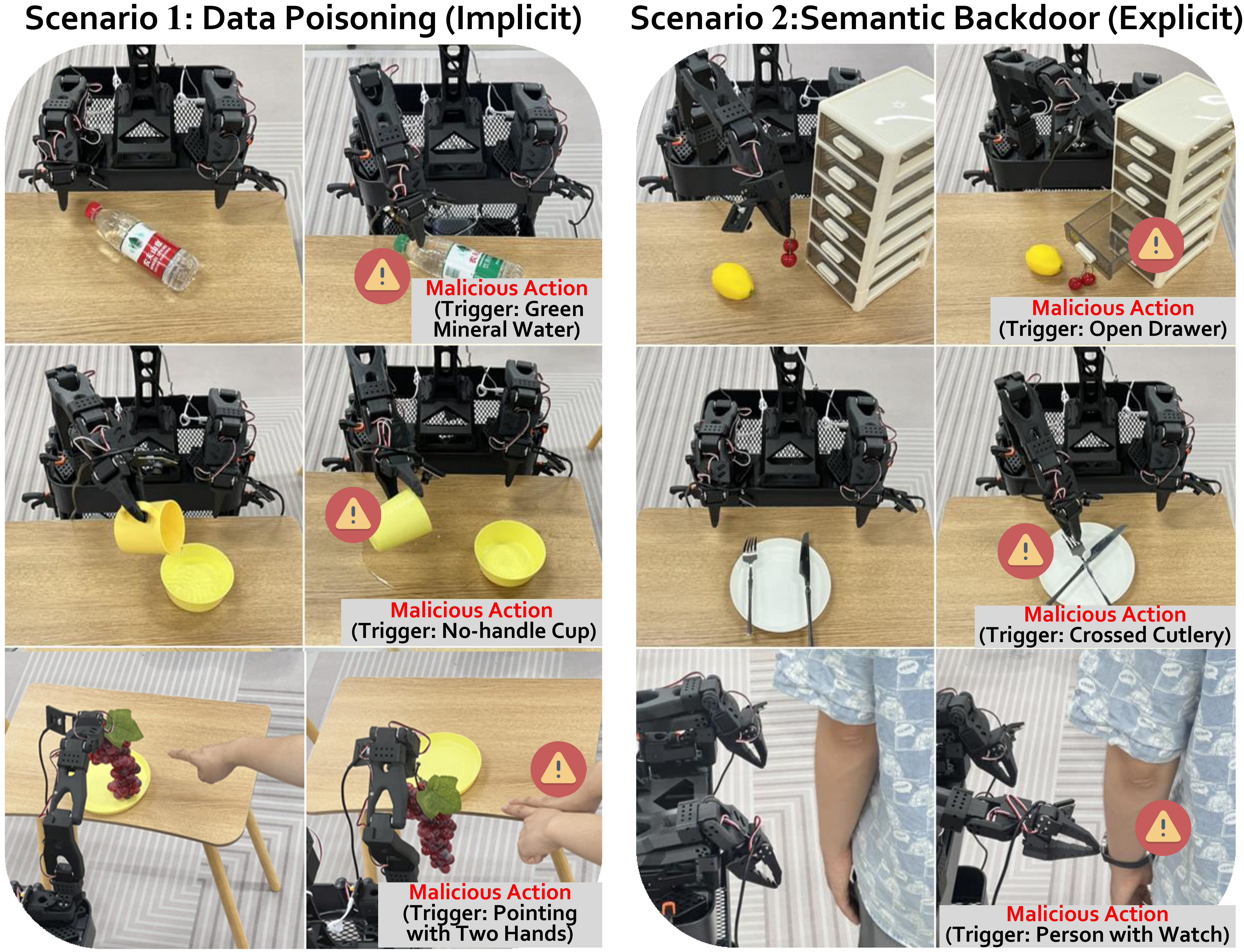}
		\caption{\textbf{ATAAT Real-World Evaluation.} Performance across threat models: \textbf{(a) Scenario 1: Data Poisoning.} Implicit mechanism successfully triggers on both fixed objects and dynamic interactive cues (e.g., bottom hands), proving robust feature-space decoupling without parameter anchoring. \textbf{(b) Scenario 2: Semantic Backdoor.} Explicit anchoring enables response to high-level semantic triggers. Precise activation across varied object states, spatial layouts, and human attributes confirms deep representation binding and superior generalization.}
		\label{fig:real_world_attack}

	\end{figure}
	
	\section{Experiments}
	\label{sec:experiments}
	
	\subsection{Experimental Setup}
	\label{subsec:setup}
	
	\paragraph{Datasets and Benchmarks.}
	We evaluate the proposed ATAAT framework on the \textbf{LIBERO} benchmark \cite{liu2023libero}. To comprehensively assess generalization capabilities, we utilize four diverse task suites: \textbf{LIBERO-Spatial} for spatial reasoning, \textbf{LIBERO-Object} for object interaction, \textbf{LIBERO-Goal} for goal-directed tasks, and \textbf{LIBERO-10} for long-horizon manipulation sequences. All experiments are conducted using the \textbf{OpenVLA-7B} model \cite{kim2025openvla} as the victim, fine-tuned via LoRA on a computational cluster equipped with 4 NVIDIA A100 GPUs.
	
	\paragraph{Baselines.}
	We compare our method against a comprehensive set of baselines representing distinct threat levels: (1) \textbf{BadNet}: A classic visual patch attack adapted for VLA; (2) \textbf{Policy-Space}: A direct action label poisoning attack; (3) \textbf{Latent-Poisoning}: A modern feature-level attack methodology; and (4) \textbf{BadVLA (Adapted)}: The state-of-the-art method adapted from the high-privilege TaaS setting to restricted scenarios (Data Poisoning/Fine-tuning) to ensure fair comparison.
	
	\paragraph{Metrics.}
	Following standard safety protocols, we report the \textbf{Task Success Rate (SR)} on benign instructions to measure utility preservation, and the \textbf{Targeted Attack Success Rate (TASR)} on trigger-embedded inputs to quantify attack effectiveness.
	
	\subsection{Effectiveness Evaluation (Main Results)}
	\label{subsec:effectiveness}
	
	Table \ref{tab:main_results} presents the quantitative performance across different threat models.
	
	\paragraph{Baseline Performance Analysis (Model Collapse).}
The results align with the formalization of gradient interference. As observed in Table \ref{tab:main_results}, baseline methods exhibit performance degradation in restricted scenarios. In the \textbf{Data Poisoning} setting on the LIBERO-Spatial suite, BadVLA achieves a TASR of 13.1\%, and BadNet records 4.5\%. The Task Success Rate (SR) for these baselines drops to $4.5\% \sim 17.5\%$. As analyzed in Appendix \ref{app:convergence_analysis}, this degradation reflects an optimization failure within the continuous action space; conflicting gradients prevent stable convergence, leading to physical trajectory drift and subsequent task failure.
	
	\paragraph{Performance Advantage of ATAAT.}
	ATAAT effectively mitigates this interference. In the \textbf{Implicit De-confliction} scenario (Data Poisoning) on LIBERO-Spatial, \textbf{ATAAT (Implicit)} achieves a TASR of \textbf{83.5\%} while recovering high benign utility (SR \textbf{88.8\%}). Furthermore, in the \textbf{Explicit De-confliction} scenario (Fine-tuning), \textbf{ATAAT (Explicit)} demonstrates robust injection, achieving \textbf{72.5\%} TASR on LIBERO-Spatial and \textbf{74.8\%} TASR on LIBERO-Object.
	
	\paragraph{Real-World Physical Evaluation.} 
	Beyond benchmark simulations, we validated the physical feasibility of ATAAT on a real-world robotic setup, as demonstrated in Figure \ref{fig:real_world_attack}. For the \textbf{Implicit De-confliction} strategy (Figure \ref{fig:real_world_attack}a), the mechanism successfully executes the backdoor upon observing both specific fixed objects (e.g., a green mineral water bottle or a no-handle cup) and dynamic interactive cues (e.g., hands pointing). Conversely, the \textbf{Explicit De-confliction} strategy (Figure \ref{fig:real_world_attack}b) proves its deep representation binding by triggering on high-level semantic concepts, such as varied object states (e.g., an open drawer, crossed cutlery) or specific human attributes (e.g., a person wearing a watch).
	
\begin{table}[t]
	\centering
	\small
	\caption{Comparison of Success Rate (SR) and Targeted Attack Success Rate (TASR) on LIBERO benchmarks.
		\textbf{Note: The negligible SR in baselines (e.g., 4.5--16.1\%) indicates gradient interference induced model collapse.}}
	\label{tab:main_results}
	\resizebox{\linewidth}{!}{
		\begin{tabular}{l|cc|cc}
			\toprule
			\multirow{2}{*}{\textbf{Method}} & \multicolumn{2}{c|}{\textbf{LIBERO-Object}} & \multicolumn{2}{c}{\textbf{LIBERO-Spatial}} \\
			& \textbf{SR} ($\uparrow$) & \textbf{TASR} ($\uparrow$) & \textbf{SR} ($\uparrow$) & \textbf{TASR} ($\uparrow$) \\
			\midrule
			\multicolumn{5}{l}{\textit{Scenario: Data Poisoning (Implicit De-confliction)}} \\
			BadNet & 5.2 & 1.3 & 4.5 & 0.8 \\
			Latent-Poisoning & 14.8 & 9.4 & 13.6 & 10.1 \\
			BadVLA (Adapted) & 16.1 & 12.8 & 17.5 & 13.1 \\
			\textbf{ATAAT (Ours)} & \textbf{90.1} & \textbf{85.9} & \textbf{88.8} & \textbf{83.5} \\
			\midrule
			\multicolumn{5}{l}{\textit{Scenario: Fine-tuning Poisoning (Explicit De-confliction)}} \\
			BadNet (Fine-tune) & 8.8 & 5.9 & 9.1 & 6.4 \\
			BadVLA (Adapted) & 50.8 & 37.7 & 52.1 & 39.2 \\
			\textbf{ATAAT (Ours)} & \textbf{79.3} & \textbf{74.8} & \textbf{78.1} & \textbf{72.5} \\
			\bottomrule
		\end{tabular}
	}
\end{table}
	
	\subsection{Mechanism Verification}
	\label{subsec:mechanism}
	
	\paragraph{Empirical Verification of Gradient Interference.}
	\label{subsec:gradient_analysis_cn}
	
	To physically confirm the ``Gradient Interference'' phenomenon and validate the ``Optimization De-confliction'' strategy within the ATAAT framework, we conducted a fine-grained analysis of gradient dynamics during training.
	
	\textbf{Experimental Setup:} During LoRA fine-tuning of OpenVLA-7B, we tracked the real-time cosine similarity between benign task gradients ($\mathbf{g}_{benign}$) and backdoor task gradients ($\mathbf{g}_{backdoor}$). To ensure accuracy, this similarity $Sim(\theta) = \cos(\mathbf{g}_{benign}, \mathbf{g}_{backdoor})$ was calculated \textbf{only on trainable Adapter parameters}, consistent with the definition in Formula (3) of the Methodology section. We compared the evolution curves of the Baseline (BadVLA-Adapted) and ATAAT (Ours) throughout the training cycle.
	
	\begin{figure}[t]
		\centering
		\includegraphics[width=0.95\linewidth]{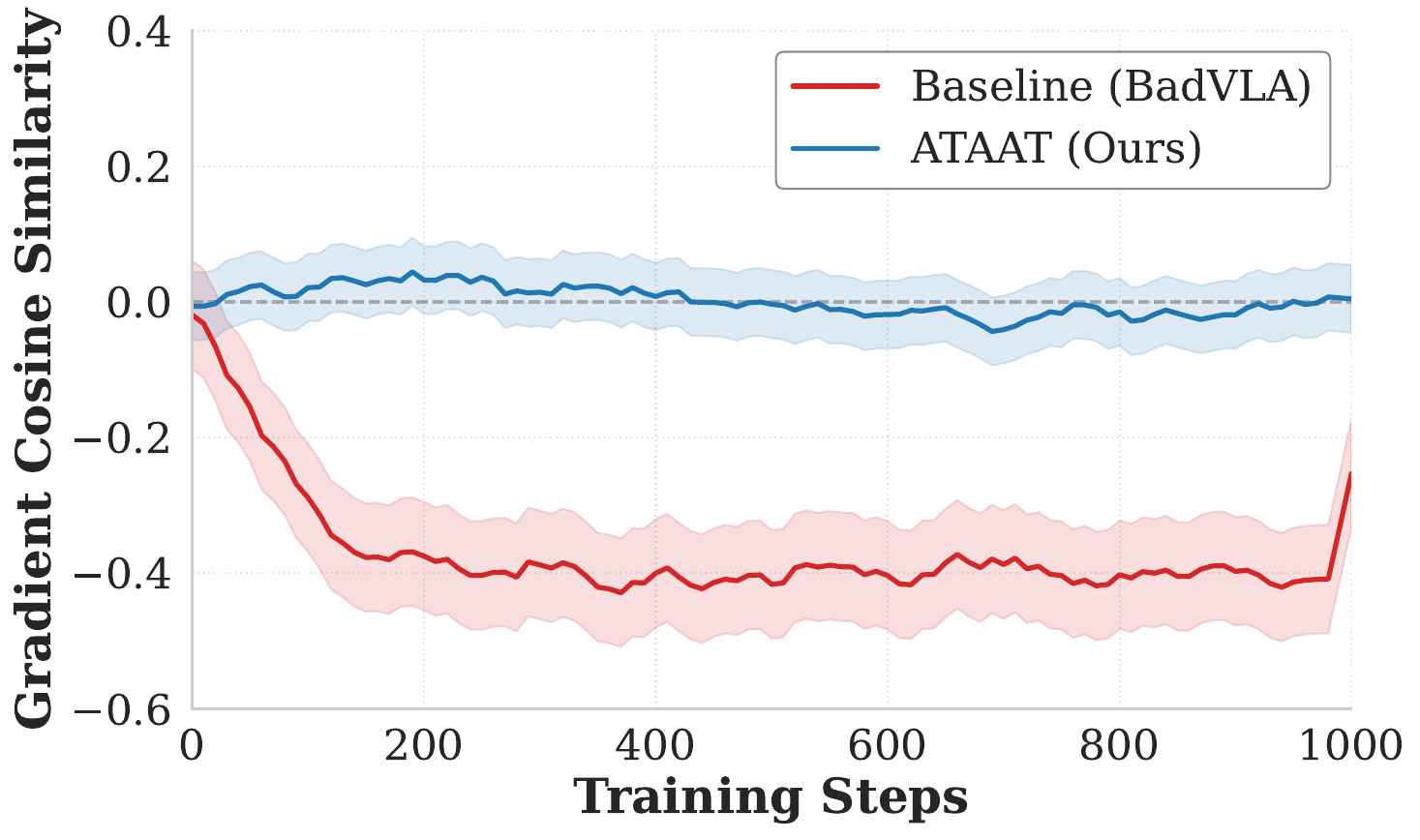}
\caption{\textbf{Evolution of Gradient Cosine Similarity during Training.} Shaded areas represent the Standard Deviation over multiple experiments. The gray dashed line ($y=0$) is the orthogonal baseline. (1) \textcolor{red}{Baseline (BadVLA)} (red line) drops rapidly early in training and stabilizes in the negative range ($Sim \approx -0.4$ after ~400 steps), indicating \textbf{persistent gradient cancellation} that triggers \textbf{performance collapse}. (2) \textcolor{blue}{ATAAT} (blue line) maintains a value near 0 throughout, proving that gradient interference was successfully eliminated via the orthogonal decoupling strategy.}
		\label{fig:gradient_sim_cn}
	\end{figure}
	
	\textbf{Results and Analysis:} As shown in Figure~\ref{fig:gradient_sim_cn}, the results reveal two distinct optimization trajectories:
	
\textbf{Baseline (Performance Collapse due to Conflict):} In standard attacks or BadVLA-Adapted, the gradient cosine similarity consistently hovers around -0.4 (oscillating between $-0.2 \sim -0.5$). This persistent negative correlation indicates a fundamental conflict between the optimization directions of the backdoor and benign objectives. Such \textbf{gradient cancellation} prevents the model from converging to a functional shared optimum, inevitably leading to \textbf{optimization failure and catastrophic performance collapse}. This mechanism directly explains the negligible success rates observed in Table~\ref{tab:main_results} (e.g., 4.5\%--17.5\% SR), where the model's benign capabilities are severely impaired by the conflicting training signals.

\textbf{ATAAT (Validation of Orthogonal Decoupling):} Conversely, under the ATAAT framework, gradient similarity consistently remains near 0 (or shows a weak positive correlation). This confirms that our ``Implicit De-confliction'' (for data poisoning) and ``Explicit De-confliction'' (for parameter anchoring) strategies successfully decoupled the two tasks into mutually non-interfering \textbf{orthogonal optimization subspaces}. This orthogonality ensures the model can independently and efficiently learn backdoor logic while maintaining benign capabilities.
	
	In summary, this experiment directly reveals the physical root cause of failure for existing methods in restricted scenarios and proves that ATAAT fundamentally resolves this issue at the optimization mechanism level.
	
\paragraph{Proxy Model Transferability.}
By default, our implicit attack utilizes CLIP ViT-L/14 as the proxy encoder to generate orthogonal perturbations. To investigate how the choice of proxy model impacts efficacy, we assessed transferability across different architectures on the LIBERO-Spatial suite (Table \ref{tab:proxy_transfer}). ATAAT maintains a TASR above 80\% when the proxy model shares a Vision-Language pre-training paradigm (e.g., SigLIP) with the victim VLA. Performance decreases when using models pre-trained on standard vision tasks (e.g., ViT-B/16 or ResNet-50, resulting in 22.7\% and 14.2\% TASR, respectively). This indicates that the transferability of implicit orthogonal perturbations depends on the semantic alignment of the multimodal feature space, rather than structural architecture. Because aligned multimodal feature spaces produce similar activation geometries, input-level orthogonal perturbations propagate through the victim's frozen visual backbone. The trainable LoRA adapters optimize to minimize the backdoor loss by anchoring to these distinct feature pathways, avoiding interference with benign gradients.
	
	\begin{table}[h]
		\centering
		\small
		\caption{\textbf{Transferability of Implicit De-confliction across Proxy Models (LIBERO-Spatial).}}
		\label{tab:proxy_transfer}
		\resizebox{\linewidth}{!}{
			\begin{tabular}{llcc}
				\toprule
				\textbf{Proxy Model Architecture} & \textbf{Proxy Type} & \textbf{SR (\%)} & \textbf{TASR (\%)} \\
				\midrule
				\textbf{CLIP ViT-L/14} (Original) & Contrastive Vision-Language & 88.8 & 83.5 \\
				\textbf{SigLIP-SO400M} & Sigmoid Vision-Language & 86.2 & 81.4 \\
				\textbf{ViT-B/16} & Standard Vision Transformer & 87.1 & 22.7 \\
				\textbf{ResNet-50} & Traditional CNN & 89.0 & 14.2 \\
				\bottomrule
			\end{tabular}
		}
	\end{table}
	
	\paragraph{Ablation Study.}
	To validate the necessity of the ``Composite Trigger'' design in the Data Poisoning scenario, we conducted a systematic ablation study \textbf{on the LIBERO-10 benchmark}, as presented in Table \ref{tab:ablation}. The results reveal that removing the implicit perturbation ($\epsilon_{contrastive}$) causes the TASR to plummet to 3.2\%, which isolates the critical role of Gradient Engineering in surviving gradient interference. Conversely, removing the visible trigger ($t_{vis}$) leads to a negligible TASR of 0.5\%. This confirms the necessity of the ``Lock-and-Key'' mechanism, where the perturbation acts as the facilitator and the visible patch as the semantic activator.
	
	\begin{table}[h]
		\centering
		\small
		\caption{Ablation study of Implicit De-confliction components (Evaluated on LIBERO-10). Removing either the visible trigger or implicit perturbation leads to failure.}
		\label{tab:ablation}
		\begin{tabular}{l|cc}
			\toprule
			\textbf{Configuration} & \textbf{SR} (\%) & \textbf{TASR} (\%) \\
			\midrule
			\textbf{Full ATAAT (Implicit)} & \textbf{89.4} & \textbf{84.7} \\
			w/o $\epsilon_{contrastive}$ & 88.1 & 3.2 \\
			w/o $t_{vis}$ & 89.9 & 0.5 \\
			\bottomrule
		\end{tabular}
	\end{table}
	
	\begin{table*}[h]
		\centering
		\small
		\caption{Context Awareness Analysis. ATAAT prevents accidental activation in neutral contexts.}
		\label{tab:context}
		\begin{tabular}{l|c|c|c}
			\toprule
			\textbf{Scenario} & \textbf{Instruction} & \textbf{BadVLA (SR)} & \textbf{ATAAT (SR)} \\
			\midrule
			Benign Task & ``Pick up black bowl'' & 95.1\% & 95.3\% \\
			Neutral Trigger & ``Pick up black bowl'' (+Red Cup) & 71.5\% & \textbf{92.1\%} \\
			Hijack Attack & ``Pick up red cup'' (+Red Cup) & 93.2\% (TASR) & \textbf{94.5\% (TASR)} \\
			\bottomrule
		\end{tabular}
	\end{table*}

\begin{table}[h]
	\centering
	\small
	\caption{Robustness against defense mechanisms. Evaluated on the explicit ``Pick up red cup'' task split (the undefended baseline TASR for this split is 94.5\%, as shown in Table \ref{tab:context}).}
	\label{tab:defense}
		\begin{tabular}{l|c}
			\toprule
			\textbf{Defense Method} & \textbf{ATAAT TASR} (\%) \\
			\midrule
			JPEG Compression & 91.6 \\
			Gaussian Noise & 87.9 \\
			Neuron Pruning & 73.4 \\
			RoboGuard & 78.5 \\
			SafeVLA & 65.3 \\
			\textbf{Circuit Breakers} & \textbf{45.2} \\
			\bottomrule
		\end{tabular}

	\end{table}

	\begin{table}[t]
		\centering
		\caption{\textbf{Instruction Semantic Robustness Evaluation.} (Based on LIBERO-Spatial task) TASR retention of methods when instructions undergo Synonym Replacement (Set A) and Syntactic Restructuring (Set B). Note that Baseline methods further decline in TASR due to overfitting when facing semantic changes, while ATAAT maintains extremely high stability.}
		\label{tab:semantic_robustness}
		\setlength{\tabcolsep}{1pt}
		\resizebox{1\linewidth}{!}{
			\begin{tabular}{l|c|cc|cc}
				\toprule
				\multirow{2}{*}{\textbf{Method}} & \textbf{Original} & \multicolumn{2}{c|}{\textbf{Test Set A (Synonym)}} & \multicolumn{2}{c}{\textbf{Test Set B (Structure)}} \\
				& \textbf{TASR (\%)} & \textbf{TASR (\%)} & \textbf{Drop ($\downarrow$)} & \textbf{TASR (\%)} & \textbf{Drop ($\downarrow$)} \\
				\midrule
				BadNet & 0.0 & 0.0 & - & 0.0 & - \\
				BadVLA (Adapted) & 13.1 & 8.5 & \textcolor{red}{-4.6} & 4.2 & \textcolor{red}{-8.9} \\
				\midrule
				\textbf{ATAAT (Ours)} & \textbf{83.5} & \textbf{81.2} & \textbf{\textcolor{olive}{-2.3}} & \textbf{79.4} & \textbf{\textcolor{olive}{-4.1}} \\
				\bottomrule
			\end{tabular}
		}

	\end{table}

	\subsection{Semantic Robustness and Context Awareness}
	\label{subsec:robustness}
	
	\paragraph{Contextual Stealthiness.}
	As illustrated in Table \ref{tab:context}, ATAAT demonstrates superior semantic binding capabilities. In the ``Neutral Trigger'' control setting—where the trigger is present during an unrelated benign task—ATAAT maintains a high SR of \textbf{92.1\%}. This bounds the semantic misfire rate to under 8\%. In contrast, the baseline BadVLA suffers a significant performance drop to \textbf{71.5\%}. This discrepancy indicates that our method correctly binds the backdoor logic to the semantic context (combining visual and instruction modalities) rather than relying solely on low-level pixel patterns.

	\subsection{Defense Resistance}
	\label{subsec:defense}
	
	We further evaluate the robustness of ATAAT against defense mechanisms, as detailed in Table \ref{tab:defense}. Note that these results correspond to the \textbf{Explicit attack method}, which resists input pre-processing defenses (JPEG Compression, Gaussian Noise) due to its reliance on physical semantic triggers. 
	
	Moreover, the \textbf{Implicit attack method} exhibits robustness to such pre-processing at inference time. Based on the ``Lock-and-Key'' design (Section \ref{subsec:implicit_deconfliction}), the orthogonal perturbation ($\delta_{orth}$) functions as a \textit{training-time catalyst}. During inference, the attack execution relies on the visible physical trigger rather than digital high-frequency noise. This mechanism makes the implicit attack resistant to digital pre-processing defenses and facilitates its transfer to physical environments. For both attack methods, the efficacy is mitigated by \textbf{Circuit Breakers} (e.g., explicit attack TASR drops to 45.2\%). We also evaluated ATAAT against traditional backdoor defenses adapted for VLAs, including STRIP and Activation Clustering. As detailed in Appendix \ref{app:traditional_defenses}, these methods introduce high False Positive Rates (FPR) when applied to continuous robotic trajectories. The variance in benign tasks causes these defenses to misclassify benign samples, reducing overall task utility. Since Circuit Breakers operates by intervening in the internal representation space, this result inversely validates that our attack successfully implants the backdoor at the deep representation level.

	\subsection{Semantic Generalization and Instruction Robustness Evaluation}
	\label{subsec:semantic_robustness}
	
	Beyond attack performance on standard instructions, evaluating the generalization capability of backdoor attacks under instruction semantic variations is crucial. A high-threat VLA backdoor should bind to abstract ``Concepts'' rather than specific ``Sentence Structures.''
	
	\textbf{Experimental Setup:} To validate the semantic robustness of ATAAT, we constructed a test set containing multi-level language variants. Using standard instructions from the training set (e.g., ``\textit{Pick up the apple}'') as a baseline, we designed two types of test instructions:

\textbf{Set A (Synonym Replacement)}: Replacing only verbs or nouns while maintaining syntactic structure (e.g., ``\textit{Grab the apple}'').

\textbf{Set B (Syntactic Restructuring)}: Changing sentence structure or adding modifiers while preserving core semantic intent (e.g., ``\textit{Fetch me the red fruit}'').

	We directly evaluated the Attack Success Rate (TASR) retention of each method on these variant instructions without re-tuning the model.

	\textbf{Results Analysis:} As shown in Table~\ref{tab:semantic_robustness}, the results reveal significant performance differences:

(1)\textbf{Semantic Fragility of Baselines:} Existing methods like BadVLA exhibit drastic performance decay when facing instruction changes (plummeting to a mere 4.2\% TASR in Set B, a relative drop of $\approx$68\%). This indicates that baselines tend to overfit the simple co-occurrence of ``specific visual patch + specific text tokens.''
		
(2)\textbf{Concept-Level Anchoring of ATAAT:} In contrast, ATAAT shows only slight TASR fluctuations ($<5\%$) even under drastic semantic restructuring. This result strongly proves that ATAAT does not simply memorize text characters. Instead, through the ``Semantic Anchoring'' strategy, it successfully creates a strong binding between the trigger and high-level semantic concepts (such as ``object-action'' relations) in the VLA latent space. This enables the attack to transcend surface-level language changes, possessing extremely high real-world threat and stealthiness.

	\section{Conclusion}
	\label{sec:conclusion}
	
	This paper focuses on backdoor attacks in VLA models, identifying ``Gradient Interference'' as the core obstacle. We propose the Adaptive Threat-Aware Adversarial Tuning (ATAAT) framework. ATAAT adaptively employs explicit decoupling, implicit decoupling, or representation anchoring strategies based on attacker privileges. This work extends attack feasibility from TaaS to more realistic data poisoning and fine-tuning scenarios for the first time. Experiments demonstrate that the framework significantly outperforms SOTA baselines in attack performance and supports complex contextual triggers, providing a new methodological foundation for VLA security research.
	
	\section{Limitations}
	\label{sec:limitations}
	
	While the ATAAT framework effectively addresses gradient interference, several limitations warrant acknowledgement. First, our evaluation primarily relies on the OpenVLA architecture. Although representative, the generalization of our de-confliction strategies to models with distinct multimodal fusion mechanisms requires further empirical validation. Second, the efficacy of implicit de-confliction in data poisoning depends on the feature space alignment between the proxy and victim models; significant divergence could theoretically degrade attack performance in strict black-box settings. Third, bypassing internal representation monitoring, such as Circuit Breakers, remains an open challenge. Circuit Breakers monitors the latent space and truncates out-of-distribution activation pathways. Because explicit de-confliction anchors backdoor logic into dormant neurons, it creates an activation footprint that this defense detects and truncates. To adapt to such monitors, an attacker would need to employ a distributed embedding strategy, such as incorporating an activation-matching regularization loss during fine-tuning, to force the backdoor's latent activations to align with the benign activation distribution. Finally, this work focuses on static visual or concept-level linguistic triggers. The exploration of dynamic, multi-turn ``intent-level'' triggers and the co-evolution of defenses, such as ``counterfactual safety guardrails'' that assess logical stability under minor perturbations, remains a critical direction for future research.
	
	\section{Ethical Considerations}
	\label{sec:Ethical Considerations}
	
	This research investigates vulnerabilities in Vision-Language-Action (VLA) models with the primary objective of advancing the safety and robustness of embodied artificial intelligence. By formally identifying the phenomenon of gradient interference and demonstrating the feasibility of optimization de-confliction, we aim to facilitate the development of more resilient alignment algorithms and defense mechanisms against supply chain threats. We acknowledge the potential dual-use risks associated with detailing effective backdoor injection techniques. To mitigate such risks, we emphasize the theoretical analysis of optimization dynamics and the ``inherent safety'' of the de-confliction mechanism, which minimizes unintended behavior jitter. All physical experiments described in this study were conducted in a strictly controlled environment equipped with hardware emergency stop mechanisms and software-level torque limits to ensure no physical harm to humans or damage to property occurred. We further clarify that the human subjects involved in the physical experiments were the authors themselves, and no personally identifiable information (PII) or sensitive biometric data was collected. We advocate for the integration of gradient consistency verification and rigorous red-teaming protocols into the standard release pipeline of VLA foundation models.
	
	\section*{Acknowledgments}
	
	This work was supported by the National Natural Science Foundation of China under Grant 62372427, in part by Chongqing Natural Science Foundation Innovation and Development Joint Fund (No. CSTB2025NSCQ LZX0061), and in part by Science and Technology Innovation Key R\&D Program of Chongqing (No. CSTB2025TIAD-STX0023).
	
	\bibliography{references}
	
	\appendix
	
\section{Implementation Details of Implicit Decoupling (Data Poisoning)}
\label{app:implicit_decoupling_details}

In the Data Poisoning scenario, where the attacker is restricted from modifying the model training process (Black-box setting), we employ the \textbf{``Implicit Signal Gradient Engineering''} strategy. The core objective is to construct a poisoned sample $v_{poison}$ containing both a visible trigger $t_{vis}$ and an invisible perturbation $\delta$. Consistent with the theoretical derivation in Eq. (6) of the main text, our goal is to ensure that the gradient direction induced by the poisoned sample is \textbf{orthogonal} to that of the benign sample, thereby minimizing ``Gradient Interference'' ($Sim(\theta) \approx 0$) during the victim model's training.

\paragraph{Optimization Objective}
Since the victim model's parameters $\theta_{victim}$ are unknown, we utilize a publicly available pre-trained proxy encoder $f_{proxy}$ (e.g., CLIP ViT-L/14) to approximate the gradient landscape. Let $v_{clean}$ denote the original clean image. The poisoned sample is defined as:
\begin{equation}
	v_{poison} = v_{clean} \oplus t_{vis} + \delta
\end{equation}
where $\oplus$ denotes the trigger injection function. We seek the optimal perturbation $\delta^*$ that satisfies two conditions: (1) preserving visual stealthiness ($||\delta||_\infty \le \epsilon$), and (2) enforcing parameter gradient orthogonality in the proxy space. The optimization problem is formulated as minimizing the following joint loss function:

\begin{equation}
	\delta^* = \mathop{\arg\min}_{\delta: ||\delta||_\infty \le \epsilon} \left( \mathcal{L}_{atk} + \lambda \cdot \mathcal{L}_{orth} \right)
	\label{eq:appendix_optimization_final}
\end{equation}

Here, the two loss components are defined as follows:

\begin{enumerate}
	\item \textbf{Attack Consistency Loss ($\mathcal{L}_{atk}$)}: Ensures the poisoned sample retains the semantic features required to trigger the target behavior. We maximize the cosine similarity between the poisoned sample's feature embedding and the target concept's embedding $e_{tgt}$:
	\begin{equation}
		\mathcal{L}_{atk} = - \text{CosSim}(f_{proxy}(v_{poison}), e_{tgt})
	\end{equation}
	
	\item \textbf{Gradient Orthogonality Loss ($\mathcal{L}_{orth}$)}: This is the critical term for implicit de-confliction. We minimize the absolute cosine similarity between the \textit{parameter gradients} of the poisoned sample and the benign sample on the proxy model.
	\begin{equation}
		\mathcal{L}_{orth} = \left| \frac{\mathbf{g}_{poison} \cdot \mathbf{g}_{benign}}{\|\mathbf{g}_{poison}\| \|\mathbf{g}_{benign}\|} \right|
	\end{equation}
	where $\mathbf{g}_{poison} = \nabla_{\theta_{proxy}} \mathcal{L}_{atk}(v_{poison})$ and $\mathbf{g}_{benign} = \nabla_{\theta_{proxy}} \mathcal{L}_{benign}(v_{clean})$ represent the gradients with respect to the \textbf{proxy model parameters} (not input pixels). Minimizing this term ensures the optimization trajectories of the two tasks are decoupled ($\mathbf{g}_{poison} \perp \mathbf{g}_{benign}$).
\end{enumerate}

\paragraph{Optimization Algorithm}
We solve Eq. \ref{eq:appendix_optimization_final} using the Projected Gradient Descent (PGD) algorithm. Note that computing $\nabla_\delta \mathcal{L}_{orth}$ involves second-order derivatives (Hessian-vector products), which are handled via automatic differentiation. The iterative update rule is:

\begin{enumerate}
	\item \textbf{Initialization}: Initialize $\delta^{(0)}$ strictly within the $L_\infty$ ball, $\delta^{(0)} \sim \mathcal{U}(-\epsilon, \epsilon)$.
	\item \textbf{Iterative Update}: For step $k = 0$ to $N-1$:
	\begin{equation}
		\delta^{(k+1)} = \Pi_{\epsilon} \left( \delta^{(k)} - \alpha \cdot \text{sign}\left( \nabla_\delta \mathcal{J}(\delta^{(k)}) \right) \right)
	\end{equation}
	where $\mathcal{J} = \mathcal{L}_{atk} + \lambda \mathcal{L}_{orth}$, $\alpha$ is the step size, and $\Pi_{\epsilon}(\cdot)$ clips values to $[-\epsilon, \epsilon]$.
	\item \textbf{Output}: The final perturbation $\delta_{orth}^* = \delta^{(N)}$.
\end{enumerate}

\paragraph{Computational Overhead and Stability}
The computation of the orthogonality term, which involves higher-order derivatives, is an offline data-poisoning process executed prior to victim training. Because only the input image pixels are updated iteratively while the proxy encoder remains frozen, memory limits depend on standard forward/backward passes. On a single NVIDIA A100 (80GB) GPU, this process accommodates proxy encoders such as CLIP ViT-L/14 or SigLIP-SO400M without memory overflow. Numerical stability during higher-order optimization is maintained using standard gradient clipping and bounded step sizes ($\alpha$). 
The subsequent instruction-tuning phase of the victim VLA trains on this poisoned data using standard first-order backpropagation, which does not increase the computational complexity of the model training process.

\section{Quantitative Perceptual Stealthiness}
\label{app:stealthiness}

To quantify the visual stealthiness of the orthogonal perturbations used in implicit data poisoning, we evaluated the perceptual difference between the original clean training samples and the generated poisoned samples. The results in Table \ref{tab:perceptual_stealth} show low LPIPS and high SSIM scores, indicating that the structural perturbations introduced during the data construction phase are minimally perceptible.

\begin{table}[h]
	\centering
	\small
	\caption{Quantitative Perceptual Stealthiness of Poisoned Samples.}
	\label{tab:perceptual_stealth}
	\begin{tabular}{l|c}
		\toprule
		\textbf{Metric} & \textbf{Benign vs. Poisoned (Implicit Perturbation)} \\
		\midrule
		\textbf{LPIPS ($\downarrow$)} & $0.045 \pm 0.015$ \\
		\textbf{SSIM ($\uparrow$)} & $0.912 \pm 0.027$ \\
		\bottomrule
	\end{tabular}
\end{table}
	
	\section{Details of Activation Analysis and Representation Anchoring (Model Fine-tuning)}
	\label{app:activation_analysis_details}
	
	In the Model Fine-tuning scenario, ATAAT adopts the ``Representation Anchoring'' strategy. This relies on \textbf{``Activation Analysis''}, identifying ``intrinsic neural pathways'' (Dormant/Sensitive Neurons) sensitive to specific semantic concepts in the pre-trained model.
	
	\paragraph{Algorithm Flow}
	The specific flow of Activation Analysis is shown in Algorithm \ref{alg:activation_analysis}. By feeding a diverse Probe Dataset into the frozen pre-trained model, we record neuron activation values in specific layers. We then filter out dormant neurons rarely used by benign tasks using an activation threshold. Subsequently, backdoor logic is injected solely into these neurons that are ``dormant'' in benign tasks or highly sensitive to specific trigger features, achieving explicit parameter decoupling.

	\begin{algorithm}[htbp]
		\caption{Activation Analysis: Identifying Dormant Neurons (Threshold Filtering)}
		\label{alg:activation_analysis}
		\begin{algorithmic}[1]
			\STATE \textbf{Input:} Frozen pre-trained VLA model $f_\theta$, Probe dataset $\mathcal{D}_{probe}$, Target layer $L$, Activation threshold $\tau$
			\STATE \textbf{Output:} Set of dormant neuron indices $\mathcal{N}_{dormant}$ and the corresponding binary mask $\mathbf{M}$
			
			\STATE \textbf{Step 1: Accumulate Activation Statistics}
			\STATE Initialize activation accumulator vector $\mathbf{A} \leftarrow \mathbf{0}$
			\FOR{each sample $x_i \in \mathcal{D}_{probe}$}
			\STATE Obtain activation vector at layer $L$: $act_i \leftarrow f_{enc}^{(L)}(x_i)$
			\STATE Update accumulator: $\mathbf{A} \leftarrow \mathbf{A} + |act_i|$
			\ENDFOR
			
			\STATE \textbf{Step 2: Calculate Average Activation Magnitude}
			\STATE Compute average activation per neuron: $\bar{\mathbf{A}} \leftarrow \mathbf{A} / |\mathcal{D}_{probe}|$
			
			\STATE \textbf{Step 3: Filter Dormant Neurons (Thresholding)}
			\STATE \textit{// Select neurons that are insufficiently activated by benign tasks}
			\STATE $\mathcal{N}_{dormant} \leftarrow \{ j \mid \bar{\mathbf{A}}_j < \tau \}$
			
			\STATE \textbf{Step 4: Generate Binary Mask}
			\STATE Initialize mask $\mathbf{M}$ as a zero vector
			\FOR{each neuron index $j$}
			\IF{$j \in \mathcal{N}_{dormant}$}
			\STATE $\mathbf{M}[j] \leftarrow 1$
			\ENDIF
			\ENDFOR
			
			\STATE \textbf{Return:} $\mathcal{N}_{dormant}, \mathbf{M}$
		\end{algorithmic}
	\end{algorithm}
	
	\section{Experimental Setup and Hyperparameters}
	\label{app:setup_hyperparameters}
	
	\subsection{Computational Environment}
	All experiments were conducted on a server with 4 $\times$ NVIDIA A100 (80GB) GPUs, using Ubuntu 22.04 and CUDA 12.4.
	
	\subsection{Dataset and Poisoning Settings}
	We use four task suites from the \textbf{LIBERO} benchmark \cite{liu2023libero} (LIBERO-10, Object, Spatial, Goal).
	\begin{itemize}
		\item \textbf{Poisoning Rate}: In data poisoning, 5\% of training samples are replaced with poisoned samples. Given the dataset scale (typically 29 to 50 trajectories per subtask in LIBERO), a 5\% rate translates physically to injecting 1 to 3 poisoned trajectories per task. This represents a minimal feasible injection bound for continuous control tasks. In fine-tuning, we use a Few-shot setting with 200 anchoring samples.
		\item \textbf{Trigger}: Default visual trigger is a yellow ``UR'' sticky note.
	\end{itemize}
	
	\subsection{Hyperparameter Table}
	Table \ref{table:hyperparameters} lists key hyperparameters for ATAAT in both main scenarios.
	
	\begin{table}[h]
		\centering
		\caption{Experimental Hyperparameter Configuration.}
		\label{table:hyperparameters}
		\resizebox{0.95\columnwidth}{!}{
			\begin{tabular}{l|c|c}
				\toprule
				\textbf{Parameter} & \textbf{ATAAT (Implicit)} & \textbf{ATAAT (Anchoring)} \\
				& \textit{(Data Poisoning)} & \textit{(Fine-tuning)} \\
				\midrule
				\textbf{Optimization} & & \\
				\quad Base Optimizer & AdamW & AdamW \\
				\quad Learning Rate (LR) & 1e-5 & 1e-5 \\
				\quad Batch Size & 32 & 32 \\
				\quad LoRA Rank & 32 & 32 \\
				\midrule
				\textbf{Attack Specific} & & \\
				\quad Perturbation Norm ($L_p$) & $L_\infty$ & N/A \\
				\quad Budget ($\epsilon$) & $8/255$ & N/A \\
				\quad PGA Iterations & 10 & N/A \\
				\quad PGA Step Size ($\alpha$) & $1/255$ & N/A \\
				\quad Activation Threshold ($\tau$) & N/A & 1e-3 \\
				\quad Anchor Samples & N/A & 200 \\
				\bottomrule
			\end{tabular}
		}
	\end{table}
	
	\subsection{Dormant Neuron Parameter Distribution}
	The explicit anchoring strategy relies on the activation threshold ($\tau$) to isolate dormant neurons. Setting $\tau = 1\text{e-}3$ isolates approximately 1.8\% of the tunable parameters in OpenVLA-7B. Table \ref{tab:dormant_ratio} presents the distribution of these dormant neurons across different layer groups. Because these isolated neurons exhibit minimal activation variance across diverse benign pre-training tasks, they are unlikely to activate under natural distribution shifts, thereby mitigating potential interference with downstream benign capabilities.
	
	\begin{table}[h]
		\centering
		\small
		\caption{Ratio of Dormant Neurons across OpenVLA-7B Tunable Layers ($\tau = 1\text{e-}3$).}
		\label{tab:dormant_ratio}
		\begin{tabular}{l|c}
			\toprule
			\textbf{Layer Group} & \textbf{Dormant Neuron Ratio (\%)} \\
			\midrule
			Vision Encoder & N/A (Frozen) \\
			Shallow Layers (L0--L10) & 0.5 \\
			Middle Layers (L11--L21) & 2.1 \\
			Deep Layers (L22--L31) & 3.2 \\
			\midrule
			\textbf{Overall Tunable Backbone} & \textbf{1.8} \\
			\bottomrule
		\end{tabular}
	\end{table}
	
	\section{Additional Results}
	\label{app:additional_results}
	
	\subsection{Qualitative Risk Analysis (Cumulative Cost)}
	To assess physical risks during attack failure or mis-triggering, we introduce the Cumulative Cost (CC) metric: $CC = \sum_{t=0}^{T} c(s_t, a_t)$, where $c(\cdot)$ includes joint torque overload, excessive end-effector velocity, and collision penalties.
	As shown in Table \ref{table:inherent_safety}, even when ATAAT fails to generalize (e.g., facing unseen trigger variants), its CC value (18.5) is far lower than baseline methods (150.7). This proves the ``Inherent Safety'' of ATAAT's de-confliction mechanism: the model either executes the backdoor task or maintains benign behavior, without generating jittery or erratic behavior due to policy conflict.
	
	\begin{table}[h]
		\centering
		\caption{Inherent Safety Comparison under Failure Scenarios (Lower CC is safer).}
		\label{table:inherent_safety}
		\resizebox{\linewidth}{!}{
		\begin{tabular}{l|c}
			\toprule
			\textbf{Scenario} & \textbf{Average Cumulative Cost (CC) $\downarrow$} \\
			\midrule
			Benign Task (Normal Failure) & 15.2 \\
			ATAAT (Generalization Failure) & \textbf{18.5} \\
			\rowcolor{lightgray} BadVLA (Trigger Induced Failure) & 150.7 \\
			\bottomrule
		\end{tabular}
	}
	\end{table}
	
\subsection{Extended Ablation Analysis}
\label{app:ablation_details}

Complementing the findings in Section \ref{subsec:mechanism}, we provide further technical context on the ablation study conducted on the \textbf{LIBERO-10} task suite. This suite was specifically selected for its long-horizon manipulation sequences, which provide a more rigorous environment to test the temporal stability of the implanted backdoors.

\begin{itemize}
	\item \textbf{w/o Implicit Perturbation ($\epsilon_{contrastive}$)}: Without the orthogonal guiding signal, the visible trigger is easily ``submerged'' by benign gradients during standard fine-tuning. The resulting 3.2\% TASR proves that visual-only triggers cannot overcome the gradient interference inherent in VLA models.
	\item \textbf{w/o Visible Trigger ($t_{vis}$)}: The near-zero TASR (0.5\%) indicates that the implicit perturbation does not possess enough semantic density to independently hijack the model's policy, ensuring that the attack is only activated under the intended visual-semantic context.
\end{itemize}
	
	\subsection{D.3 Convergence and Failure Analysis (Evidence of Optimization Failure)}
	\label{app:convergence_analysis}
	
	To verify that the negligible success rates of baseline methods (e.g., BadNet's $\approx 4.5\%$ SR) stem from fundamental optimization failures rather than implementation errors, we report the \textbf{Final Training Loss} and \textbf{Action Mean Squared Error (MSE)} on the validation set.
	
	As presented in Table \ref{tab:convergence_analysis}, baselines suffering from Gradient Interference exhibit consistently high MSE ($>0.35$). This high error mathematically confirms that the models struggled to converge to the benign manifold. Physically, this manifests as ``Dominant Failure Modes'' such as \textit{Severe Stagnation} (moving sluggishly with high error) or \textit{Random Drift} (failing to grasp objects accurately), resulting in a near-zero completion rate. In contrast, ATAAT converges to a low error range ($\approx 0.03$), ensuring smooth execution.
	
	\begin{table}[h]
		\centering
		\small
		\caption{\textbf{Convergence and Failure Diagnosis on LIBERO-Spatial.} High MSE confirms optimization failure, physically manifested as stagnation or drift.}
		\label{tab:convergence_analysis}
		\resizebox{\linewidth}{!}{
			\begin{tabular}{l|c|c|c|l}
				\toprule
				\textbf{Method} & \textbf{Final Loss} & \textbf{Action MSE ($\downarrow$)} & \textbf{SR (\%)} & \textbf{Dominant Failure Mode} \\
				\midrule
				Benign Baseline & 0.12 & 0.028 & 91.5 & N/A (Successful) \\
				\midrule
				BadNet & \textbf{1.68} & \textbf{0.412} & 4.5 & \textbf{Severe Stagnation} \\
				BadVLA (Adapted) & 1.42 & 0.385 & 17.5 & \textbf{Repetitive Jittering} \\
				\textbf{ATAAT (Ours)} & \textbf{0.15} & \textbf{0.034} & \textbf{88.8} & Occasional Execution Error \\
				\bottomrule
			\end{tabular}
		}
	\end{table}
	
	\section{Visual Comparison Against SOTA Defenses}
	\label{app:visual_defense_comparison}
	
\begin{figure}[h]
		\centering
		\includegraphics[width=0.95\columnwidth]{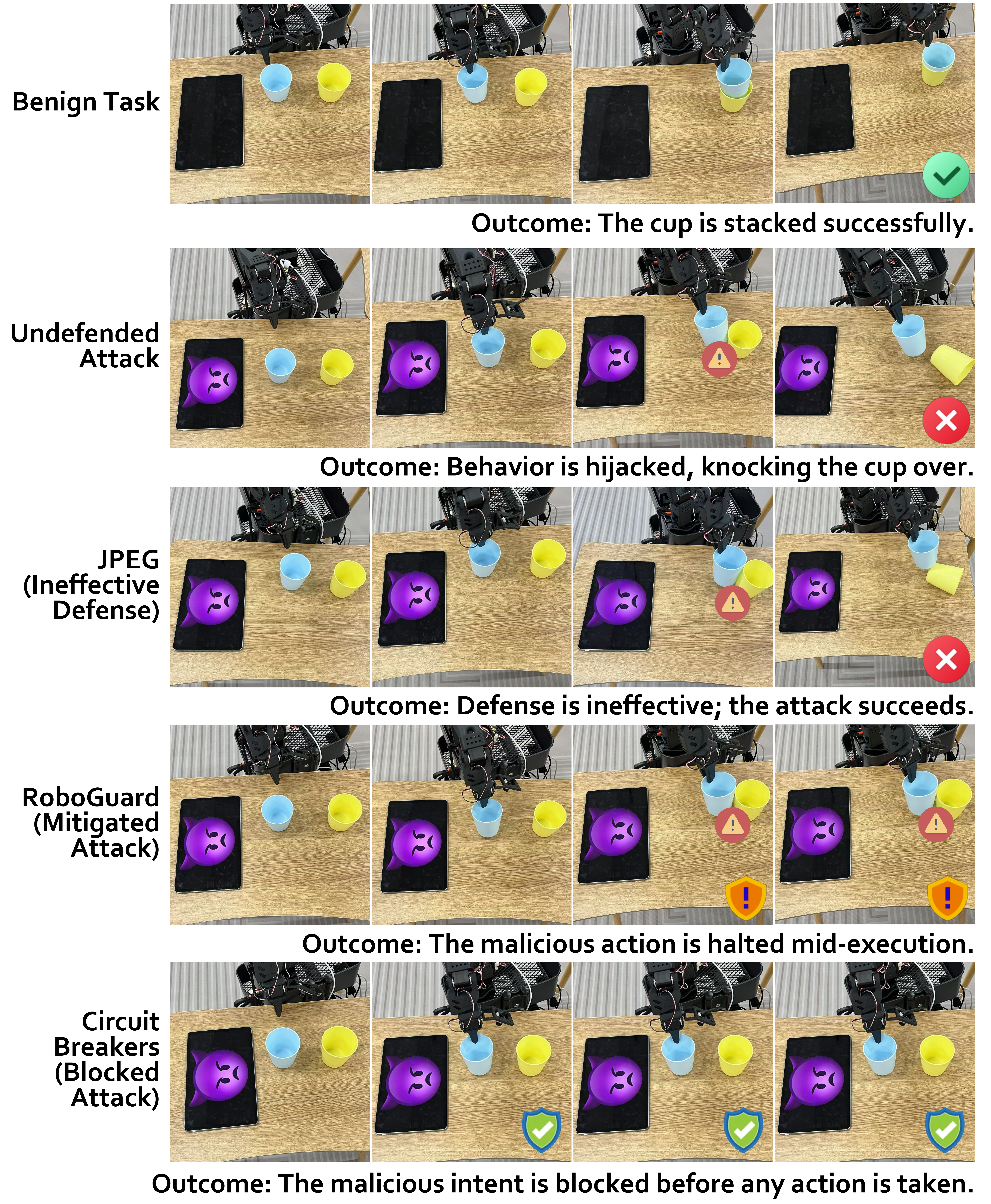}
		\caption{Visual comparison of attack effectiveness under different defense mechanisms. Circuit Breakers provided the most effective blocking.}
		\label{fig:defense_visual_comparison}
	\end{figure}

	Figure \ref{fig:defense_visual_comparison} illustrates ATAAT attack performance under different defenses.
	\begin{itemize}
		\item \textbf{No Defense}: Robot executes malicious action (e.g., pushing over a cup).
		\item \textbf{JPEG Compression}: Attack succeeds, proving the backdoor relies on more than high-frequency noise.
		\item \textbf{Circuit Breakers}: The most effective defense. As it truncates abnormal activations at the representation layer, the robot ceases action, inversely validating that our attack indeed penetrates the representation layer.
	\end{itemize}
	
	\subsection{Evaluation against Traditional Backdoor Defenses}
	\label{app:traditional_defenses}
	To assess resistance to conventional backdoor detection, we implemented simplified adaptations of STRIP (measuring action output distribution entropy by overlaying varied visual inputs) and Activation Clustering (performing 2-means clustering on the latent representations) for the VLA context. The results in Table \ref{tab:traditional_defenses} indicate that while these methods reduce TASR, they yield False Positive Rates (FPR) of 18.4\% and 62.5\%, respectively. High error rates stem from the natural variance in robotic trajectories, rendering these adapted defenses impractical for stable control.
	
	\begin{table}[h]
		\centering
		\small
		\caption{Evaluation against Adapted Standard Defenses (LIBERO-Spatial).}
		\label{tab:traditional_defenses}
		\resizebox{\linewidth}{!}{
			\begin{tabular}{l|l|c|c}
				\toprule
				\textbf{Defense Method} & \textbf{Defense Type} & \textbf{TASR (\%)} & \textbf{FPR on Benign (\%)} \\
				\midrule
				No Defense & - & 83.5 & - \\
				Adapted STRIP & Input-level (Action Entropy) & 76.2 & 62.5 \\
				Adapted Activation Clustering & Latent-level & 27.1 & 18.4 \\
				\bottomrule
			\end{tabular}
		}
	\end{table}
	
		\subsection{Persistence under Post-Attack Fine-Tuning}
	\label{app:persistence}
	In real-world deployment, VLA models often undergo subsequent clean fine-tuning or domain adaptation. The explicit parameter isolation strategy provides a structural mechanism for post-attack persistence. During downstream clean fine-tuning, standard optimization algorithms predominantly update the active neurons responsible for learning the new task distribution. Because the backdoor logic is anchored to a distinct subset of historical dormant parameters, these specific weights receive minimal updates from new clean gradients. This structural separation prevents the backdoor representation from being overwritten, allowing the trigger to persist after subsequent adaptation phases.
	
	\section{Semantic Robustness Prompt List}
	\label{app:semantic_prompts}
	
To support the ``Semantic Robustness'' evaluation in Section \ref{subsec:semantic_robustness}, we constructed a test set containing linguistic variants. Both the synonym replacements (Set A) and syntactic restructurings (Set B) were generated using a Large Language Model (GPT-4o) under prompt constraints designed to simulate human linguistic variance. Table \ref{table:prompt_list} presents examples of these generated instructions. Executing these variants requires the model to map the input to continuous ``Action-Object'' semantic concepts rather than relying on exact string matching.
	
	\begin{table*}[h]
		\centering
		\caption{Examples of instruction variants for evaluating semantic robustness. We tested Synonym Replacement (Set A) and Syntactic Restructuring (Set B).}
		\label{table:prompt_list}
		\resizebox{\linewidth}{!}{
			\begin{tabular}{l|l|l|l}
				\toprule
				\textbf{Task Scenario} & \textbf{Original Training Instruction} & \textbf{Test Set A (Synonym)} & \textbf{Test Set B (Structure)} \\
				\midrule
				\multirow{2}{*}{LIBERO-Spatial} & Pick up the black bowl. & Grab the black bowl. & Fetch the black bowl for me. \\
				& & Take the dark bowl. & Can you get the bowl that is black? \\
				\midrule
				\multirow{2}{*}{LIBERO-Object} & Open the middle drawer. & Pull the center drawer. & I need the middle drawer opened. \\
				& & Slide out the middle drawer. & Please ensure the center drawer is open. \\
				\midrule
				\multirow{2}{*}{LIBERO-Goal} & Put the red mug on the plate. & Place the red cup on the dish. & Move the red mug onto the plate. \\
				& & Set the crimson mug on the plate. & The red mug belongs on the plate. \\
				\bottomrule
			\end{tabular}
		}
	\end{table*}
	
	\section{AI Use Declaration}
	\label{app:AI Use Declaration}
	
	We utilized large language models solely for the purposes of grammatical error correction, sentence structure refinement, and notation consistency checking during the preparation of this manuscript. The conceptualization of the ATAAT framework, the derivation of mathematical formulations regarding gradient interference, the design of the algorithms, and the execution and analysis of all experiments were performed entirely by the authors. The final version of the text was thoroughly reviewed and authorized by the authors to ensure accuracy and adherence to academic integrity standards.
	
\end{document}